\documentclass[lettersize,journal]{IEEEtran}
\usepackage{amsmath,amsfonts}
\usepackage{array}
\usepackage[caption=false,font=normalsize,labelfont=sf,textfont=sf]{subfig}
\usepackage{textcomp}
\usepackage{stfloats}
\usepackage{url}
\usepackage{verbatim}
\usepackage{graphicx}
\def\BibTeX{{\rm B\kern-.05em{\sc i\kern-.025em b}\kern-.08em
    T\kern-.1667em\lower.7ex\hbox{E}\kern-.125emX}}
\usepackage{balance}
\usepackage{multirow}
\usepackage{booktabs}
\usepackage{algorithm}
\usepackage{algpseudocode}
\usepackage{makecell} 

\begin{document}

\title{TemporalGS: Training-Free Plug-and-Play Acceleration for 3D Gaussian Splatting Rendering via Temporal Priors}
\author{Yuhongze Zhou, Zihao Yang, Xinxin Zuo, Juwei Lu
\thanks{Y. Zhou is with McGill University, Montr\'{e}al, Canada. (email: yuhongze.zhou@mail.mcgill.ca)}
\thanks{Z. Yang is with University of Waterloo, Waterloo, Canada. (email: zh2yang@uwaterloo.ca)}
\thanks{X. Zuo is with Concordia University, Montr\'{e}al, Canada. (email: xinxinzuo2353@gmail.com)}
\thanks{J. Lu is with University of Toronto, Toronto, Canada. (email: juwei@ieee.org)}
}

\markboth{Journal of \LaTeX\ Class Files,~Vol.~18, No.~9, September~2020}%
{How to Use the IEEEtran \LaTeX \ Templates}

\maketitle

\begin{abstract}
3D Gaussian Splatting (3DGS) has revolutionized novel-view synthesis with its fast and high-fidelity rendering. However, rendering at high FPS and low latency across various scenes remains a challenge, especially when large amounts of 3D Gaussian ellipsoids appear in the scene. To address this issue, we introduce TemporalGS, to the best of our knowledge, the first training-free plug-and-play algorithmic approach to accelerate 3DGS rendering without any post-training or post-processing, implemented on top of tile-based software rasterization. The key idea is that, instead of rendering frames independently as 3DGS, we leverage the temporal priors, represented by novel geometry and appearance buffers, etc., to reduce redundancy of Gaussian preprocessing, sorting, and rasterization operations of consecutive frames. Specifically, we propose two acceleration strategies: (1) temporal dynamic culling, which filters out Gaussians that contribute less to current frame rendering; (2) selective rendering, which renders only a small portion of tiles that cannot be approximated by the temporal priors. By adapting and interleaving these two strategies, TemporalGS yields a simple but effective plug-and-play solution for 3DGS rendering speed-up without any training. Extensive experiments show that TemporalGS achieves comparable or even better performance compared to existing state-of-the-art post-training or post-processing-based 3DGS rendering acceleration approaches. TemporalGS can significantly enhance the rendering speed of various 3DGS methods, achieving up to $1.48\times$ acceleration, while maintaining competitive rendering quality. We further extend our TemporalGS to hardware rasterization-based 3DGS to show the portability of our algorithm.

\end{abstract}

\begin{IEEEkeywords}
3D Gaussian Splatting, Rendering acceleration.
\end{IEEEkeywords}

\section{Introduction}
\label{sec:intro}

\IEEEPARstart{3}{D} Gaussian Splatting (3DGS)~\cite{kerbl20233d} has recently emerged as a more powerful 3D scene representation than Neural Radiance Field (NeRF) for novel-view synthesis with efficient training and rendering. 3DGS utilizes tile-based software rasterization and consists of preprocessing, sorting, and rasterization steps. Among these three steps, sorting and rasterization are the two major 3DGS computational burdens that correlate with the number of Gaussians. To further enhance the user roaming experience, this paper focuses on enhancing the rendering efficiency of 3DGS with higher FPS and lower latency across various-scale 3D scenes, especially large-scale scenes.

Existing 3DGS rendering acceleration methods focus mainly on two aspects, including Gaussian pruning/culling~\cite{fan2024lightgaussian, papantonakis2024reducing,lee2024compact,hanson2025speedy,niemeyer2025radsplat,liu2025occlugaussian} and implementation optimization~\cite{lee2024gscore,feng2025flashgs,liao2025tc}. For pruning/culling, many works, such as LightGaussian~\cite{fan2024lightgaussian} and C3DGS~\cite{lee2024compact}, prune redundant Gaussians during (post-)training to improve rendering speed. RadSplat~\cite{niemeyer2025radsplat} and OccluGaussian~\cite{liu2025occlugaussian}, to speed up test-time rendering, introduce visibility-based filtering strategies as post-processing to eliminate Gaussians that are unlikely to contribute to rendering for each viewpoint. GScore~\cite{lee2024gscore} and FlashGS~\cite{feng2025flashgs} introduce more precise Gaussian-tile intersection tests to reduce the number of Gaussians online. For implementation optimization, GScore~\cite{lee2024gscore} presents a hardware acceleration unit for 3DGS on the mobile GPU. FlashGS~\cite{feng2025flashgs} introduces an optimized execution pipeline, refined control and scheduling mechanisms, and memory access optimizations. TC-GS~\cite{liao2025tc} introduces a hardware-dependent module utilizing Tensor Cores for 3DGS speed-up. In addition, StochasticSplats~\cite{kheradmand2025stochasticsplats} introduces a sorting-free 3DGS by post-training, which uses the number of Monte Carlo samples to trade off computation time and quality.

\begin{figure}[tp]
\setlength\tabcolsep{0pt}
\centering
\resizebox{0.95\columnwidth}{!}{
\begin{tabular}{cc}
(a) 3DGS & (b) TemporalGS \\
\includegraphics[width=0.5\columnwidth]{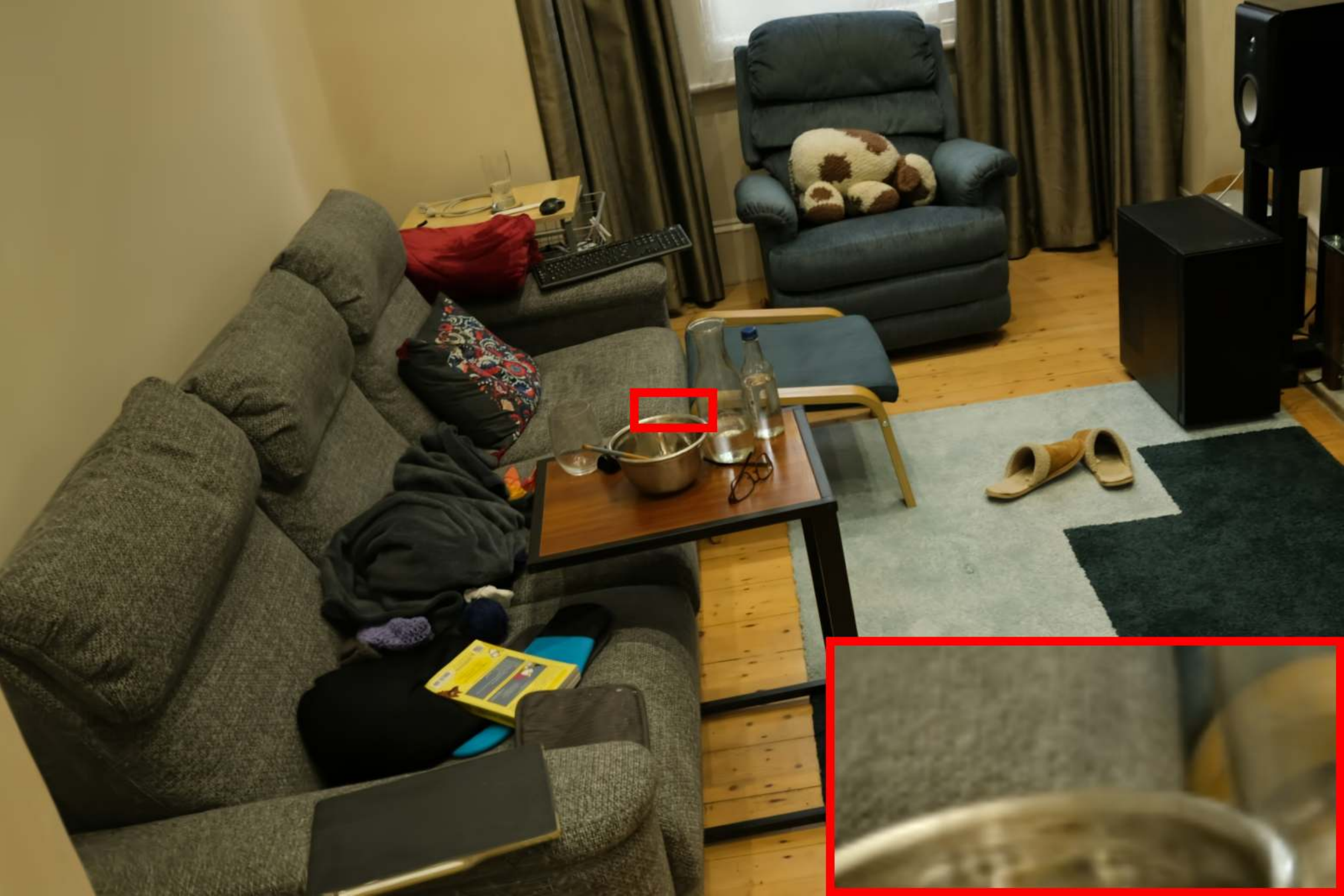}&\includegraphics[width=0.5\columnwidth]{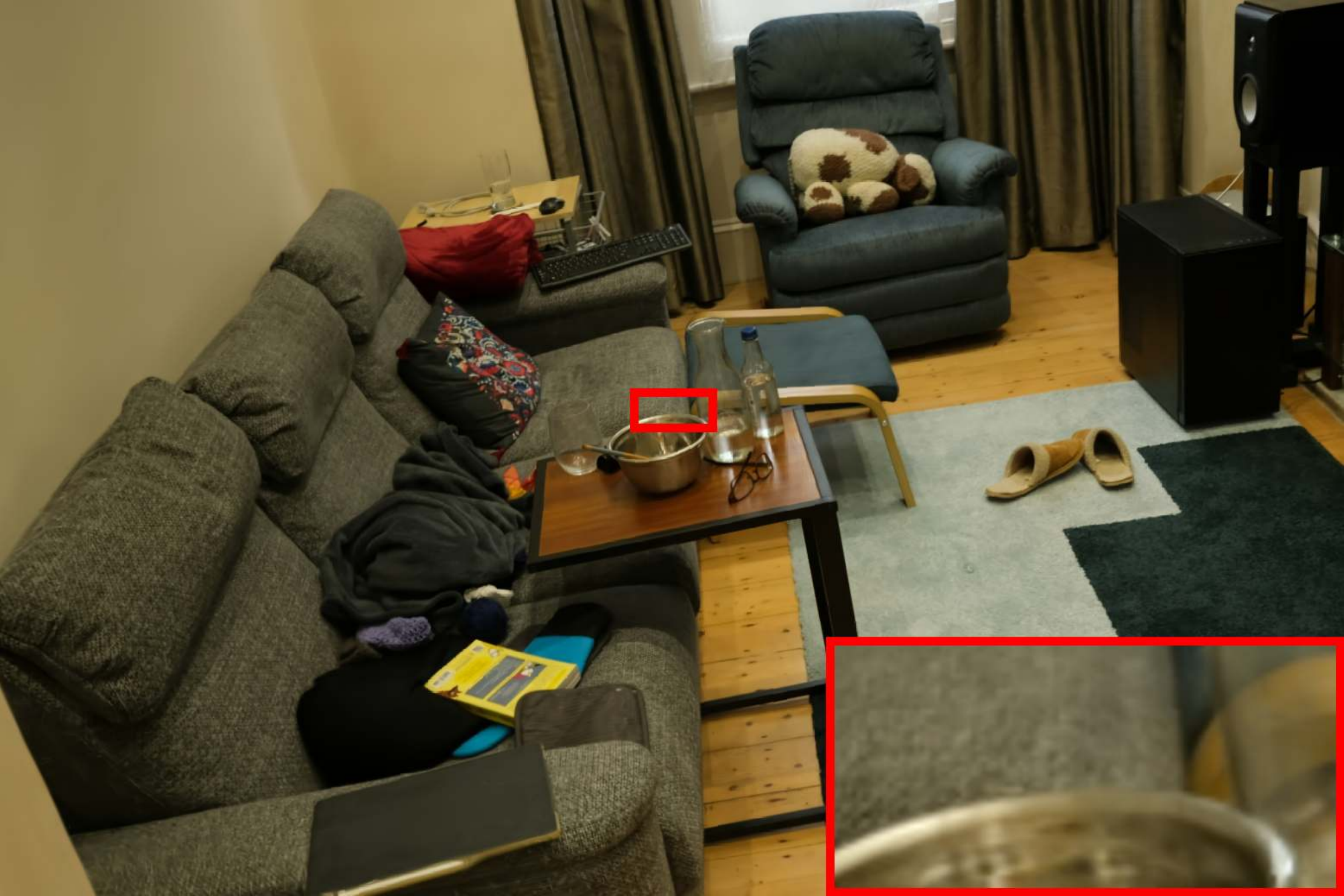}\\
FPS: 259 & FPS: 383\\
\includegraphics[width=0.5\columnwidth]{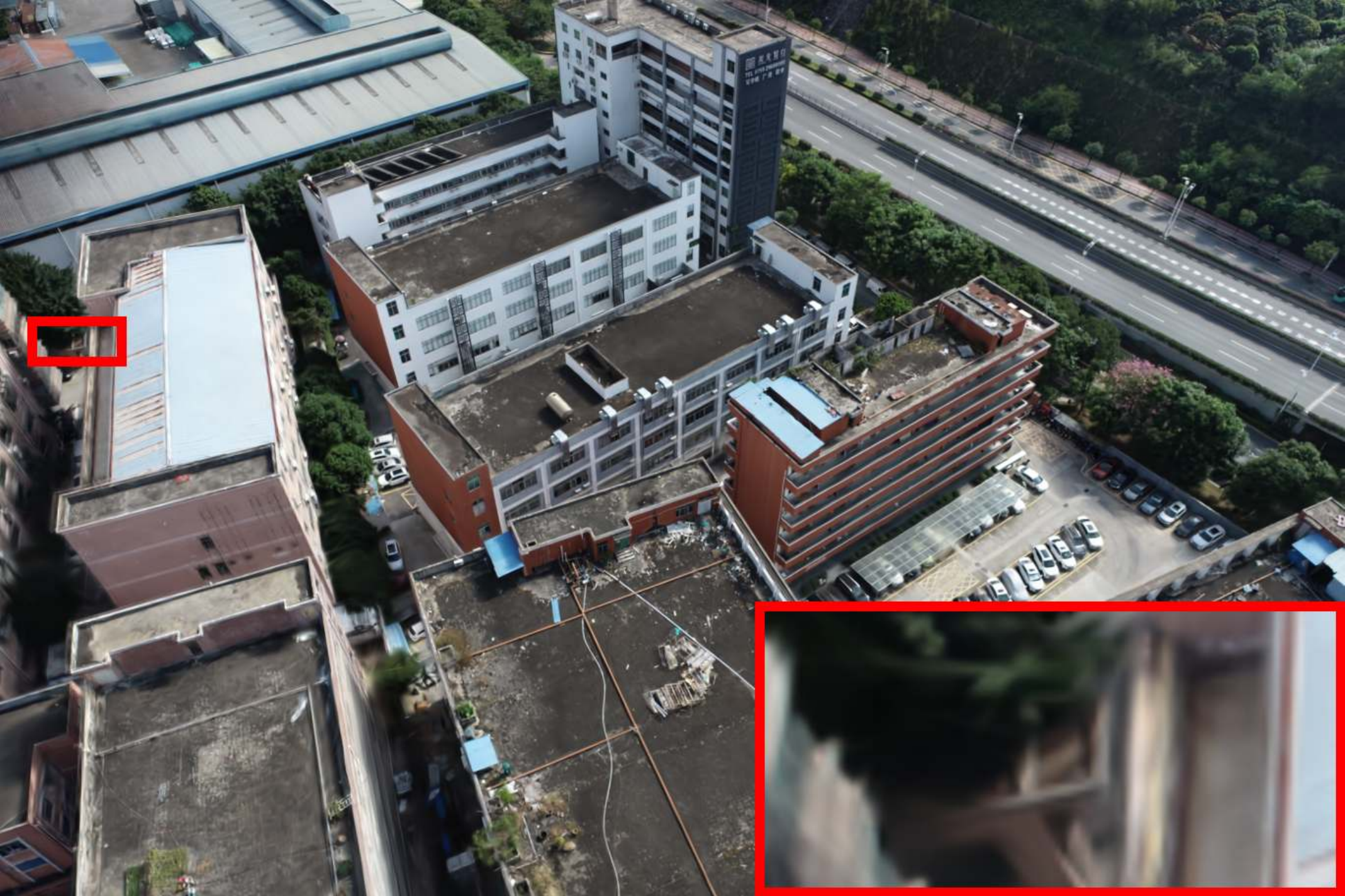}&\includegraphics[width=0.5\columnwidth]{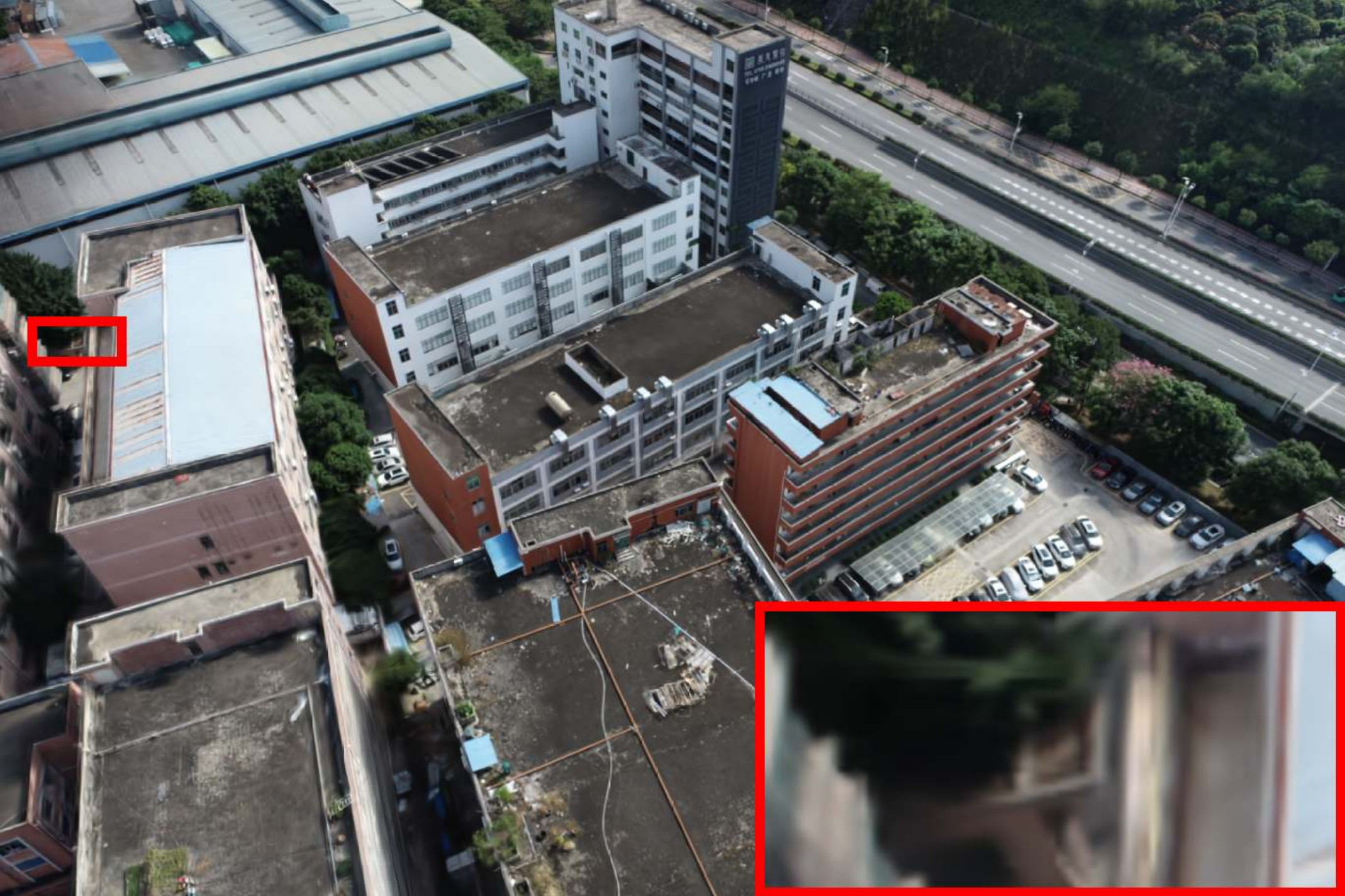}\\
FPS: 126 & FPS: 320
\end{tabular}}
\caption{Our TemporalGS can speed up rendering across various-scale scenes and maintain similar visual quality as 3DGS.}
\label{fig:teaser}
\end{figure} 

In contrast to existing 3DGS rendering acceleration approaches that rely on (post-)training/post-processing, or are tightly coupled with the 3DGS rendering pipeline and the underlying hardware, we propose TemporalGS, the first training-free plug-and-play algorithmic approach, to speed up 3DGS rendering without any post-training or post-processing. Our key motivation is that, when rendering neighboring frames, the sets of 3D Gaussians contributing to the rendering exist significant overlap. The rendering of a previous frame actually provides valuable information to reduce the redundancy of Gaussian preprocessing, sorting, and rasterization operations of the current frame rendering. Specifically, we propose novel temporal priors of the previous frame, which contain temporal geometry and appearance buffers, etc. We further introduce two strategies to strain the capability of temporal priors for efficient new frame rendering. First, we introduce temporal dynamic culling to eliminate unnecessary Gaussians. It contains two culling policies: (1) adaptive frustum culling to filter Gaussians that are physically outside the new viewpoint; (2) temporal occlusion culling to filter Gaussians behind the warped temporal geometry buffer (TGB). Second, we propose a selective rendering strategy to reduce the sorting and rasterization cost. It leverages a tile selection algorithm to only render a small portion of tiles that cannot be approximated by the warped temporal appearance buffer (TAB). Then, we construct new frames by stitching rendered tiles and the warped TAB together. During the stream rendering process, our TemporalGS combines and interleaves the above two strategies to achieve high FPS and low latency rendering. Through extensive experiments, we demonstrate that training-free TemporalGS is able to achieve performance on par with, or surpassing, existing state-of-the-art post-training or post-processing-based 3DGS rendering acceleration methods. In addition, TemporalGS can be seamlessly incorporated as a plug-in module into software- or hardware-based 3DGS rasterization pipelines for rendering acceleration, while preserving competitive image quality, as illustrated in Fig.~\ref{fig:teaser}.

Our contributions can be summarized as follows: \textbf{(1)} We introduce TemporalGS, the first training-free plug-and-play algorithm, to improve the rendering efficiency of 3DGS or its variants up to 1.48$\times$ without any post-training or post-processing. \textbf{(2)} With our novel temporal prior representations, our temporal dynamic culling and selective rendering strategies efficiently reduce the number of Gaussians and select a small set of tiles to render online. \textbf{(3)} Our training-free TemporalGS achieves comparable or better performance than the state-of-the-art post-training or post-processing-based 3DGS rendering acceleration methods.

\section{Related Work}
\label{sec:related_work}

\subsection{3D Gaussian Splatting}
Despite the superiority of 3DGS in 3D reconstruction and novel view synthesis, there remain issues that hinder its real application, such as excessive and incontrollable memory consumption, difficult training of large-scale scenes, and its inefficient design and implementation. Therefore, there are three main streams of work that address each issue, respectively.

To reduce memory consumption, many attempts~\cite{papantonakis2024reducing,fan2024lightgaussian,mallick2024taming,lee2024compact,niedermayr2024compressed,hanson2025speedy,niemeyer2025radsplat} have been made. They usually leverage pruning and vector quantization or knowledge distillation to compress Gaussian features during training for smaller model size and faster training and rendering. To improve 3DGS training on large-scale scenes, previous work leverages the divide-and-conquer approach~\cite{lin2024vastgaussian,liu2024citygaussian,kerbl2024hierarchical,liu2025occlugaussian} or introduces a 3DGS distributed training system~\cite{chen2024dogs,zhao2025on,li2024retinags}.

Instead of addressing the first two issues, some papers~\cite{mallick2024taming,durvasula2023distwar,lee2024gscore,feng2025flashgs,hanson2025speedy,niemeyer2025radsplat,liao2025tc} focus on improving 3DGS algorithm design and implementation in training or rendering. For training optimization, Taming3DGS~\cite{mallick2024taming} derives faster and numerically equivalent solutions for gradient computation and parameter updates, leading to efficient backpropagation. DISTWAR~\cite{durvasula2023distwar} is a software approach to accelerate atomic operations during gradient computation. For rendering optimization, GScore~\cite{lee2024gscore} tailors the 3DGS implementation to the mobile GPU with Gaussian shape-aware intersection test, hierarchical sorting, and sub-tile skipping. FlashGS~\cite{feng2025flashgs} applies algebraic simplifications to its precise Gaussian-tile intersection in preprocessing and introduces an optimized pipeline to enhance hardware utilization. TC-GS~\cite{liao2025tc} is hardware-dependent and relies on mapping alpha-blending to matrix multiplications to fully utilize Tensor Cores for rendering acceleration. RadSplat~\cite{niemeyer2025radsplat} clusters input cameras and gathers visible Gaussians for each region to reduce redundancy. CityGaussian~\cite{liu2024citygaussian} generates different detail levels using LightGaussian~\cite{fan2024lightgaussian} and introduces an adaptive block-wise LOD rendering strategy. OccluGaussian~\cite{liu2025occlugaussian} introduces a region-based rendering strategy that culls invisible Gaussians by region-based visibility calculation and subdivision. Our TemporalGS acts as a training-free plug-and-play algorithmic approach to speed up 3DGS and is complementary to these methods.

\subsection{Temporal Coherence based Rendering}
Temporal coherence has been widely investigated to accelerate surface-based rendering~\cite{shade1998layered,bowles2012iterative,hauswiesner2013temporal,muller2021real,xu2022temporal,yang2024mob,wu2024gffe}, volume rendering~\cite{elek2012interactive,iglesias2020real}, point cloud visualization~\cite{schutz2020progressive}, and NeRF rendering~\cite{li2023steernerf,liu2024nerfbuff,steiner2024frustum}. However, it is infeasible to directly adapt these acceleration methods to 3DGS. Also, recent concurrent works~\cite{oh2026neo,tao2025gs} show some efforts to accelerate 3DGS with temporal coherence. NeoG~\cite{oh2026neo} proposes a reuse-and-update sorting algorithm, but is only applicable to sorting-based Gaussian Splatting and provides relatively limited benefits when implemented purely in software. GS-Cache~\cite{tao2025gs} eliminates redundant anchor decoding of Scaffold-GS based variants across consecutive frames and introduces a system-level framework utilizing multi-GPU parallel rendering for large-scale scenes on VR devices. Different from existing works, TemporalGS fully explores temporal coherence in 3DGS rendering and introduces a training-free plug-and-play acceleration solution for both software- and hardware-based 3DGS rasterization.

\begin{figure*}[tp]
    \centering
     \includegraphics[width=\textwidth]{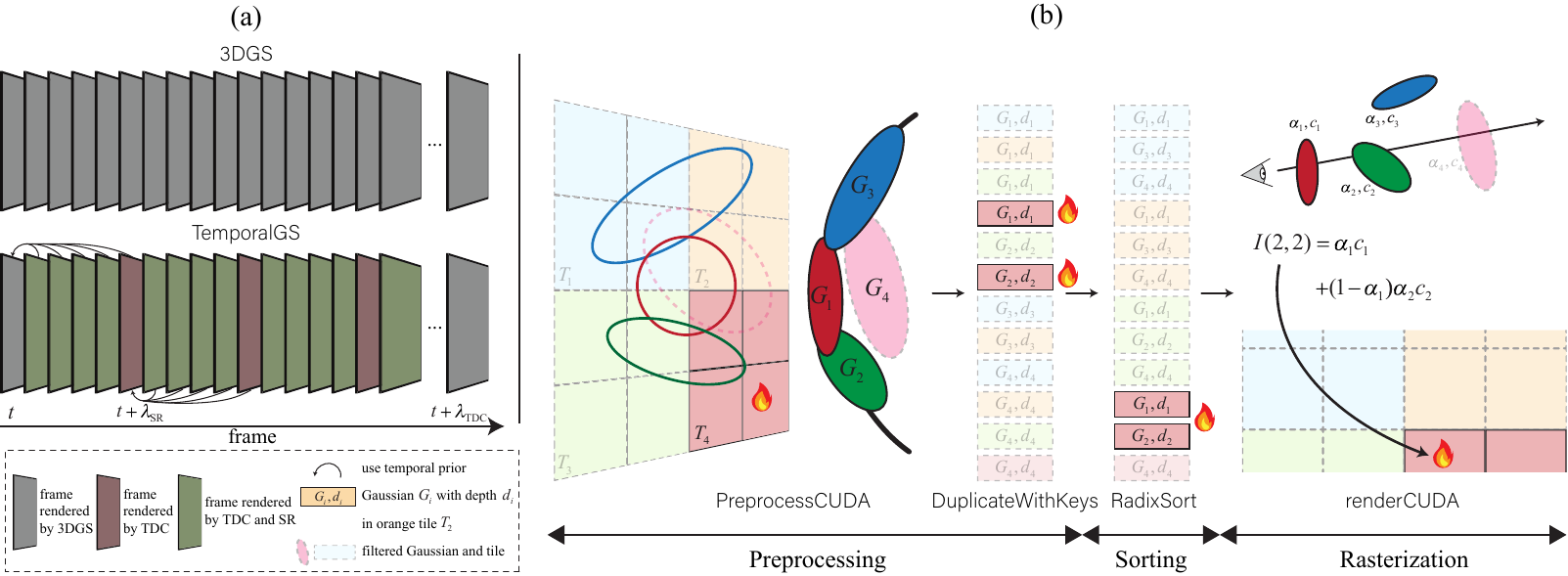}
\caption{(a) We present the high-level comparison about rendering a sequence of frames between vanilla 3DGS and TemporalGS. TemporalGS leverages temporal priors for efficient rendering while 3DGS renders independently. (b) We illustrate how our TemporalGS speeds up 3DGS forward pass. Please refer to Section~\ref{subsec:preliminary} for 3DGS forward pass details. TemporalGS leverages temporal dynamic culling to filter a redundant Gaussian $G_{4}$ behind the surface in the preprocessing step. It uses selective rendering to only render a tile $T_{4}$ and approximates $T_1$, $T_2$, and $T_3$ by the warped temporal appearance buffer. Therefore, it significantly reduces sorting and rasterization cost. Unnecessary Gaussians and tiles are marked by dotted line.}
    \label{fig:temporalgs}
\end{figure*}

\section{Approach}
\label{sec:approach}
\subsection{Preliminary: 3D Gaussian Splatting}
\label{subsec:preliminary}
3D Gaussian Splatting (3DGS)~\cite{kerbl20233d} models a scene with a set of 3D
Gaussians $\mathbf{G}$. Each 3D Gaussian $G_i \in \mathbf{G}$ is parameterized by mean $\mu_i$, covariance matrix $\Sigma_i$, opacity $o_i$, and spherical harmonics (SH) coefficients. It can be divided into three steps: preprocessing, sorting, and rasterization. In the preprocessing step, given a world-to-camera transformation matrix $W$, intrinsic matrix $K$, and a 16$\times$16 tile-split screen space $\mathbf{T}$, near-plane culling is performed to filter Gaussians behind the camera. Each remaining Gaussian is transformed and projected onto the 2D screen space. Gaussians that do not intersect with any tiles are discarded, while Gaussians that intersect with multiple tiles are duplicated. Then, each Gaussian is characterized by the 2D mean $\mu_i^{\prime}$, 2D covariance matrix $\Sigma_i^{\prime}$, opacity $o_i$, depth $d_i$, and color $c_i$. The sorting process, for each tile $T_k \in \mathbf{T}$, sorts its intersected Gaussians by depth from front to back. For simplicity, Gaussians that intersect with $T_k$ are denoted as $\{G_i \mid G_i \in T_k\}$. Finally, the rasterization step renders an image using $\alpha$-blending.

\subsection{Problem Formulation and Overview}
We formulate our problem as follows. We perform two types of rendering, i.e., vanilla 3DGS rendering and TemporalGS accelerated rendering. A frame rendered by the former is considered as a reference frame. Our goal is to accelerate the rendering of $\lambda-1$ frames $\{\bar{I}_{t+1}, \bar{I}_{t+2}, \dots \bar{I}_{t+\lambda-1}\}$ between two neighboring reference frames $I_{t}$ and $I_{t+\lambda}$.

In Fig.~\ref{fig:temporalgs}, we provide an overview of our TemporalGS. As shown in Fig.~\ref{fig:temporalgs}(a), Unlike 3DGS which renders each frame independently, we utilize temporal priors to accelerate new frame rendering. Specifically, when rendering a reference frame, we construct the temporal priors by caching temporal geometry and appearance buffers, etc. Then, we introduce two strategies, i.e., temporal dynamic culling (TDC) and selective rendering (SR). They collaborate and interleave with each other adaptively between every $\lambda_\textup{TDC}$ and $\lambda_\textup{SR}$ interval to take full advantage of temporal priors for efficient rendering.

In the following subsections, we first describe the detailed procedure for constructing temporal priors (Section~\ref{subsec:priors}) and then present the two key strategies that utilize those temporal priors for rendering acceleration (Section~\ref{subsec:tdc}, Section~\ref{subsec:sr}).

\subsection{Construction of Temporal Priors}
\label{subsec:priors}
\noindent\textbf{Temporal Geometry Buffer (TGB)}
Given the scene geometry seen from the reference frame, we create a temporal geometry buffer (TGB) as a geometric prior. Assuming that neighboring frames usually share a similar geometry, we can propagate the TGB to subsequent frames through a warping operation.

In detail, while rendering the reference frame $I_t$, we can compute its frame depth at pixel $(x,y)$ as
\begin{equation}
    D_t(x,y)=\frac{\sum_{G_i\in T_k}{w_i(x,y) d_i}}{\sum_{G_i\in T_k}w_i(x,y)},
\end{equation}
and its standard deviation as
\begin{equation}
    \sigma_t(x,y)=\sqrt{\frac{\sum_{G_i\in T_k}{w_i(x,y)(d_i - D_t(x,y))^2}}{\sum_{G_i\in T_k}w_i(x,y)}},
\end{equation}
where the pixel $(x,y)$ lies within the tile $T_k$. $\{D_t, \sigma_t\}$ forms the temporal geometry buffer (TGB).

At frame $t+i$, $i\in [1,\lambda-1]$, given $D_t$, $\sigma_t$, the world-to-camera transformation matrices of the reference frame and frame $t+i$, denoted as $W_{t}$ and $W_{t+i}$, and the intrinsic matrix $K$, we can map the pixel $(x,y)$ in the reference frame to frame $t+i$ by forward warping:
\begin{equation}
    (x^\prime, y^\prime) \simeq KW_{t+i} W_{t}^{-1} D_t(x,y) K^{-1}[x,y,1]^T,
    \label{eq:foward_depth_warping}
\end{equation}
where we ignore the last element of homogeneous coordinates for illustration simplicity. According to this, we can not only do depth warping but also unproject the depth standard deviation map $\sigma_t$ using $D_t$ to 3D space and reproject them to frame $t+i$. The warped destination usually lands ``between'' pixels, so we clip it to its nearest pixel. Since multiple source pixels might overlap at the same destination, we preserve the geometry closer to the camera and the maximum depth standard deviation to obtain high recall. Note that before warping, the default warped depth and standard deviation are initialized by $z_\textup{far}+\epsilon$ and 0, respectively. We denote the warped $D_t$ and $\sigma_t$ at frame $t+i$ as $\overrightarrow{D}_{t\rightarrow {t+i}}^{\min}$ and $\overrightarrow{\sigma}_{t\rightarrow {t+i}}^{\max}$, which maintain the minimum projected depth and the maximum depth standard deviation at the destination.

After obtaining the warped TGB $\{\overrightarrow{D}_{t\rightarrow {t+i}}^{\min}, \overrightarrow{\sigma}_{t\rightarrow {t+i}}^{\max}\}$, we can model the surface range that appears in the reference frame as $[\overrightarrow{D}_{t\rightarrow {t+i}}^{\min} -\lambda_1 \overrightarrow{\sigma}_{t\rightarrow {t+i}}^{\max}, \overrightarrow{D}_{t\rightarrow {t+i}}^{\min} + \lambda_2 \overrightarrow{\sigma}_{t\rightarrow {t+i}}^{\max}]$ at frame $t+i$, where $\lambda_1$ and $\lambda_2$ are set to adjust the variance range. We find that this simple surface formulation works well for 3DGS. In our experiments, we set $\lambda_1$ and $\lambda_2$ to 3 and 0.

\begin{figure*}[tp]
    \centering
     \includegraphics[width=0.9\textwidth]{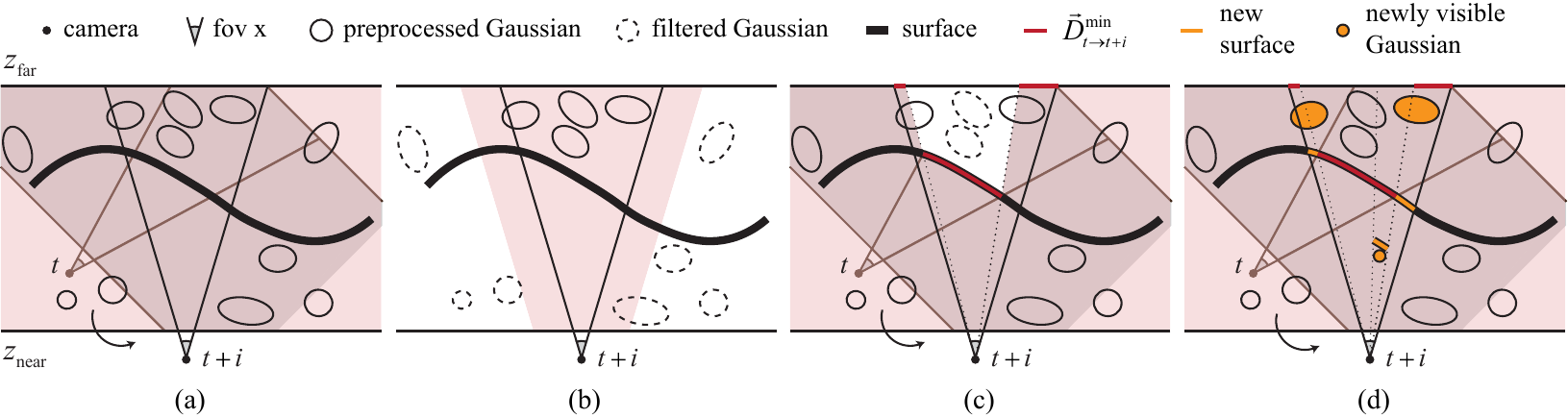}
    \caption{We visualize, at frame $t+i$, (a) 3DGS near-plane culling, (b) adaptive frustum culling, (c) temporal occlusion culling, and (d) selective rendering regions. All Gaussians within the pink region are preprocessed at frame $t+i$.}
    \label{fig:temporalgs_details}
\end{figure*}

\noindent\textbf{Temporal Appearance Buffer (TAB)}
Since consecutive frames are usually similar, we can use the reference frame $I_t$, known as the temporal appearance buffer, to approximate the shared content of the frame $I_{t+i}$. Given TAB and the warped depth $\overrightarrow{D}_{t\rightarrow {t+i}}^{\min}$, we can obtain the warped TAB $\overleftarrow{I}_{t\rightarrow {t+i}}$ by backward warping for a fast rendering approximation. Here we use bilinear interpolation.

\noindent\textbf{Temporal Gaussian Buffer}
Thanks to temporal coherence, the Gaussians that contribute to the rendering of $I_t$ are likely to contribute to the rendering of frame $t+i$. We define this set of Gaussians as
\begin{math}
\mathbf{G}_{\textup{rendered}} = \{G_i \mid \max\limits_{(x,y)\in I_t}(w_i(x,y))>0\}
\end{math}, and consider it as the temporal Gaussian buffer.

\subsection{Temporal Dynamic Culling (TDC)}
\label{subsec:tdc}
Given the temporal priors of the reference frame $I_t$, at frame $t+i$, $i\in [1,\lambda_\textup{TDC}-1]$, we perform the following two dynamic cullings to reduce the number of preprocessed Gaussians.

\noindent\textbf{Adaptive Frustum Culling (AFC)}
As shown in Fig.~\ref{fig:temporalgs_details}(a), since vanilla 3DGS performs near-plane culling, it can result in the computational cost of projecting redundant Gaussians onto 2D screen. Although a regular frustum constrained by the field of view has been widely used in traditional rendering, simply applying it to 3DGS leads to artifacts around image boundaries. Therefore, for the rendering of $I_t$, we define an adaptive frustum as $F_{\min} = \min\limits_{G_i\in \mathbf{G}_{\textup{rendered}}}(ndc_i)$ and $F_{\max} = \max\limits_{G_i\in \mathbf{G}_{\textup{rendered}}}(ndc_i)$, where $ndc_i = \textup{Pix2NDC}(\mu_i^\prime)$ is the normalized device coordinate (NDC) of Gaussian $G_i$ and the function $\textup{Pix2NDC}(\cdot)$ transforms the screen space coordinates to NDC coordinates. To render frame $t+i$, after near-plane culling, we only keep Gaussians within our adaptive frustum, denoted as $\{G_i \mid F_{\min}\leq ndc_i \leq F_{\max}\}$, for further preprocessing.

\noindent\textbf{Temporal Occlusion Culling (TOC)}
The Gaussians that are far behind the visible surface are usually unnecessary for rendering. To this end, during the rendering of frame $t+i$, we perform occlusion culling on Gaussians whose 2D means are within the screen space to remove Gaussians behind the warped TGB. The filtered Gaussians are denoted as
\begin{math}
    \{G_i \mid G_i \notin \mathbf{G}_\textup{rendered} \text{ and } d_i > \overrightarrow{D}_{t\rightarrow {t+i}}^{\min}(\bar{\mu}_i^\prime) + \lambda_2 \overrightarrow{\sigma}_{t\rightarrow {t+i}}^{\max}(\bar{\mu}_i^\prime) \}
\end{math}, where $\bar{\mu}_i^\prime$ is the nearest screen pixel position of Gaussian $G_i$. Note that in $\overrightarrow{D}_{t\rightarrow {t+i}}^{\min}$ and $\overrightarrow{\sigma}_{t\rightarrow {t+i}}^{\max}$, pixels that do not have warped geometry maintain $z_\textup{far}+\epsilon$ as depth and 0 as standard deviation, as shown in Fig.~\ref{fig:temporalgs_details}(c). This mechanism helps to include Gaussians that represent a new surface closer to the camera.

\subsection{Selective Rendering (SR)}
\label{subsec:sr}
The content of most regions and their view-dependent effects are similar among neighboring frames. As a result, only those dissimilar regions need to be rendered at frame $t + i$, while others can be efficiently warped from the previous reference frame. As shown in Fig.~\ref{fig:temporalgs_details}(d), regions that appear different and cannot be warped due to geometrical discrepancy between frame $t$ and $t+i$ include: (1) the occluded regions, where Gaussians that do not contribute to $I_t$ form a new surface closer to the camera in frame $t+i$; (2) the unmatched regions, where it fails to establish pixel-wise geometrical correspondences with the reference frame. Since 3DGS utilizes tile-based software rasterization, we name the tiles in dissimilar regions as ``selected tiles'' $\mathbf{T}_{\textup{selected}}$, which should go through the forward rendering pass. We detect those tiles using the following processes.

\noindent\textbf{Occluded Tile Detection} To render $\bar{I}_{t+i}$ properly, we need to detect tiles that are occluded by a new surface. As shown in Fig.~\ref{fig:temporalgs_details}(d), we first detect newly visible Gaussians, 
\begin{math}
    \{G_i \mid G_i \notin \mathbf{G}_\textup{rendered} \text{ and } d_i \leq \overrightarrow{D}_{t\rightarrow {t+i}}^{\min}(\bar{\mu}_i^\prime) - \lambda_1 \overrightarrow{\sigma}_{t\rightarrow {t+i}}^{\max}(\bar{\mu}_i^\prime) \}
\end{math}, which do not contribute to $I_t$ and are closer to the camera than ${G}_\textup{rendered}$. To avoid false positives, we filter out the newly visible Gaussians with trivial contribution to their intersected tiles. In our work, we use $\alpha$ to measure the contribution and set the filtering threshold to be $\frac{1}{255}$. We denote the tiles selected by the occlusion detection process as $\mathbf{T}_\textup{selected}^{\textup{oc}}$. 

\noindent\textbf{Unmatched Tile Detection}
Due to viewpoint changes, some regions of the depth map $\overrightarrow{D}_{t\rightarrow {t+i}}^{\min}$ may not have warped geometry, i.e., having $z_\textup{far}+\epsilon$ value. We consider those regions as no-hit regions where pixel-to-pixel matching with the reference frame can fail. To take full advantage of TAB, we propose a depth filling method to repair $\overrightarrow{D}_{t\rightarrow {t+i}}^{\min}$. For each no-hit pixel, if all its $3\times 3$ neighboring pixels have valid values, we fill this pixel with the average of its neighboring depths. We denote the repaired depth as $\overrightarrow{D}_{t\rightarrow {t+i}}^{\min\prime}$. Based on the repaired depth, we perform the pixel-to-pixel correspondence query by a backward version of Equation~\eqref{eq:foward_depth_warping} to check if all pixels within a tile can find their correspondence in the reference frame. If not, the tile is selected for rendering and is denoted as $\mathbf{T}_\textup{selected}^{\textup{um}}$. Note that the obtained pixel-to-pixel correspondence information is also used for later backward warping.

\noindent\textbf{Adaptive Warping and Rendering}
Combining $\mathbf{T}_\textup{selected}^{\textup{oc}} \cup \mathbf{T}_\textup{selected}^{\textup{um}}$, we can obtain the tiles that need to be rendered at frame $t+i$, denoted as $\mathbf{T}_\textup{selected}$. Then, the Gaussians intersected with $\mathbf{T}_\textup{selected}$ are sorted, and the frame $t+i$ is rendered as
\begin{equation}
    \bar{I}_{t+i} (x,y) =
    \begin{cases}
I_{t+i}(x,y),   \quad (x,y)\in \mathbf{T}_{\textup{selected}}, \\ 
\overleftarrow{I}_{t\rightarrow {t+i}}^{\prime}(x,y), \quad (x,y)\in \mathbf{T}\setminus \mathbf{T}_{\textup{selected}}, 
\end{cases},
\end{equation}
where $\overleftarrow{I}_{t\rightarrow {t+i}}^{\prime}$ is the warped TAB based on the repaired depth $\overrightarrow{D}_{t\rightarrow {t+i}}^{\min\prime}$. We find that this direct stitching does not produce noticeable sewn lines and much view-dependent discrepancies, as shown in Fig.~\ref{fig:teaser}. Please see more visual results in the supplement. 
In real applications, we can encounter significant camera motion and the less accurate surface representation of 3DGS. It can increase unmatched tiles, lead to severe warping errors, and decrease TemporalGS speed gain. Therefore, for a trade-off of rendering efficiency and quality, we propose an adaptive reference frame insertion mechanism. Specifically, we introduce pseudo reference frames, which are standard reference frames with AFC. The adaptive mechanism forces the frame $t+i+1$ to be a pseudo reference frame if the number of selected tiles at frame $t+i$ exceeds $\eta (\%)$ of the total tiles. We find that $\eta=40\%$ works well in the work.

\section{Experiments}
\subsection{Benchmark}
\noindent\textbf{Datasets} (1) Small-scale scene datasets: We evaluate our approach on nine scenes of Mip-NeRF-360~\cite{Barron2021MipNeRF3U}. Following the common protocol, images are downsampled by a factor of 4 for outdoor scenes and by a factor of 2 for indoor scenes. (2) Large-scale aerial and street-view scene datasets: For aerial views, we not only adopt Small City scene of synthetic MatrixCity~\cite{li2023matrixcity} dataset, but also use public real-world scene datasets, including Residence, Building, and Rubble~\cite{turki2022mega}. We use the same image resolution configuration as CityGaussian~\cite{liu2024citygaussian}. For street views, we evaluate our method on Small City scene of Hierarchical Gaussian~\cite{kerbl2024hierarchical} dataset. (3) Large-scale indoor scene datasets: We choose OccluScene3D~\cite{liu2025occlugaussian}, which contains canteen, class building, and gallery scenes, for its complex layouts and significant occlusions. Note that we provide quantitative results and analysis of OccluScene3D in the supplement.

\noindent\textbf{Smooth Trajectory Data Generation}
We not only synthesize smooth trajectories, but also use real user roaming trajectories to evaluate performance in a real environment. For synthetic trajectories, we fit curves based on given consecutive test camera poses and interpolate camera poses along curves at a fixed interval. For real trajectories, we record user movements in 3DGS SIBR viewer.

\noindent\textbf{Metrics} We compute PSNR, SSIM, and LPIPS on originally provided test images. We report the averaged frames per second (FPS) for all images along the trajectories.

\begin{table*}[th!]
\setlength{\tabcolsep}{2pt}
\huge
\caption{Quantitative comparison on five large-scale aerial and street-view scene datasets. Mem means Peak Memory (GB).}
\label{tab:large_scale}
\centering
\resizebox{\textwidth}{!}{
\begin{tabular}{l|l|cccc|cccc|cccc|cccc|cccc|ccccc}
\toprule
 \multirow{2}{*}{Category} & \multirow{2}{*}{Methods}& \multicolumn{4}{c|}{MatrixCity} &\multicolumn{4}{c|}{Residence}&\multicolumn{4}{c|}{Building}&\multicolumn{4}{c|}{Rubble} &\multicolumn{4}{c|}{Small City} & \multicolumn{5}{c}{Average} \\
&& PSNR$\uparrow$ & SSIM$\uparrow$ & LPIPS$\downarrow$ & FPS$\uparrow$& PSNR$\uparrow$ & SSIM$\uparrow$ & LPIPS$\downarrow$ & FPS$\uparrow$& PSNR$\uparrow$ & SSIM$\uparrow$ & LPIPS$\downarrow$ & FPS$\uparrow$ & PSNR$\uparrow$ & SSIM$\uparrow$ & LPIPS$\downarrow$ & FPS$\uparrow$ & PSNR$\uparrow$ & SSIM$\uparrow$ & LPIPS$\downarrow$ & FPS$\uparrow$ & 
Mem & PSNR$\uparrow$ & SSIM$\uparrow$ & LPIPS$\downarrow$ & FPS$\uparrow$\\
\midrule
Baseline & CityGS  & 27.44 &	0.866 & 0.205 & 106 &22.03 & 0.814 & 0.213 & 126 & 21.96 & 0.784 & 0.243 &	95 & 25.59	& 0.811	& 0.233	& 195 & 25.13 & 0.802	& 0.295	& 90 & 8.22 & 24.43 & 0.815 & 0.238 &  122\\ 
\midrule
\multirow{3}{*}{Post-training} & LightGS & 26.54 & 0.824 & 0.283 & 291& 21.57 &0.793&0.232 & 238 & 21.56 & 0.751 & 0.279 & 268 & 	24.65 & 0.747 & 0.296&	347 & 19.55 & 0.649 & 0.463 & 190 &  2.23 & 22.77 & 0.753& 0.311 & 267\\
& CP3DGS& 25.86 & 0.806 & 0.276 & 153 & 21.19	& 0.758	& 0.270	& 175& 20.34 & 0.693 & 0.332 & 159 & 24.00 & 0.731	&	0.314&	251  & 24.89 & 0.775 & 0.330 & 140 & 2.50 & 23.26 & 0.753 & 0.304 & 176 \\
& StochasticSplats & 24.73 & 0.654 & 0.420 & 101 & 20.15 & 0.566 & 0.418 & 84 & 19.18	& 0.504 & 0.431	& 90 & 21.63 & 0.522 & 0.406 & 128 & 23.24 & 0.538	& 0.456	& 62 & 7.44 & 21.79 & 0.557 & 0.426 & 93 \\
\midrule 
Training & C3DGS& 23.05 &	0.666 & 0.475 &	405& 19.84 &0.708&0.335&	220 & 18.44 & 0.596 & 0.444 & 375 &22.19 &  0.674	& 0.384	 & 423 & 22.89 & 0.722 & 0.417 & 898 & 1.27 & 21.28 & 0.673 & 0.411 & 464 \\
\midrule
Post-processing & RadSplat$^{\dagger}$ & 26.14 & 0.848 & 0.221 & 181 & 21.92 & 0.813	& 0.214 & 192 & 21.88 & 0.783 & 0.244 &	181 & 25.55	& 0.810	& 0.234	& 274 & 25.13 & 0.802 & 0.295 & 250 & 8.28 & 24.12 & 0.811 & 0.241 & 216\\
\midrule
Temporal-coherence & NeoG & 27.14 &	0.864 & 0.206 & 131 & 21.95 & 0.814	& 0.213	& 174 & 21.55 & 0.781	& 0.246	& 72 & 25.26 & 0.810 & 0.233 & 242 & 21.29 & 0.765 & 0.322 & 33 & 12.37 & 23.44 & 0.807 & 0.244 & 130
\\
\midrule
\multirow{6}{*}{\makecell[l]{Training-free\\plug-and-play}} & CityGS-TGS & 26.57 & 0.834 & 0.219 & 236 & 21.16 & 0.758 & 0.237 & 320 & 21.41 & 0.738 & 0.261 & 271 & 24.66 & 0.752	&0.255	& 409  & 24.67	& 0.787	& 0.303	& 200 & 8.48 & 23.69 & 0.774 & 0.255 & 287\\
& LightGS-TGS& 25.85 & 0.795 & 0.293 & 540 & 20.90	& 0.743	& 0.255 & 550 & 21.12 & 0.710 & 0.293 & 609 & 24.15	& 0.705	& 0.309	& 753  & 19.59	& 0.646	& 0.465	& 345 & 2.35 & 22.32 & 0.720 &  0.323 & 559 \\
& CP3DGS-TGS& 	25.37 & 0.784	& 0.288	& 340 & 20.61 & 0.715 & 0.290 & 476 & 19.92	& 0.652	& 0.346 & 455 & 23.46 & 0.684 & 0.332 & 602  & 24.54 & 0.763 & 0.337 & 310 & 2.72 & 22.78 & 0.720 & 0.319 & 437 \\
& StochasticSplats-TGS & 24.74 & 0.670 & 0.397 & 146 & 20.23 & 0.594 & 0.386 & 169 & 	19.34 &	0.521 & 0.407 &	200 & 22.16	& 0.556	& 0.373	& 259 & 23.46 & 0.581 & 0.430 & 161 & 7.70 & 21.99 & 0.584 & 0.399 & 187 \\
& C3DGS-TGS& 22.97 & 0.660 & 0.477 & 668 & 19.51 & 0.677	& 0.348 & 633 & 18.32 & 0.574 & 0.451 & 977 & 22.00 & 0.647 & 0.393 & 973 &	22.65 &  0.714 & 0.420 & 1337 & 1.31 & 21.09 & 0.654 & 0.418 & 918\\
& RadSplat$^{\dagger}$-TGS & 25.41 & 0.817 & 0.234 & 320 & 21.06 & 0.755 & 0.239 & 385 & 21.32	& 0.735 & 0.262	& 376 & 24.61 & 0.750 & 0.256 & 498 & 24.62	& 0.785	& 0.304	& 381 & 8.54 & 23.40 & 0.769 & 0.259 & 392\\
\bottomrule
\end{tabular}}
\end{table*}

\begin{table*}[th!]
\setlength{\tabcolsep}{2pt}
\huge
\caption{Quantitative comparison between software and hardware rasterization on Mip-NeRF-360 and five large-scale aerial and street-view scene datasets. web-splat failed on MatrixCity that has over 20 million Gaussians.}
\label{tab:hardware}
\centering
\resizebox{\textwidth}{!}{
\begin{tabular}{l|cccc|cccc|cccc|cccc|cccc|cccc}
\toprule
 \multirow{2}{*}{Methods}& \multicolumn{4}{c|}{Mip-NeRF-360} & \multicolumn{4}{c|}{MatrixCity} &\multicolumn{4}{c|}{Residence}&\multicolumn{4}{c|}{Building}&\multicolumn{4}{c|}{Rubble} &\multicolumn{4}{c}{Small City} \\
& PSNR$\uparrow$ & SSIM$\uparrow$ & LPIPS$\downarrow$ & FPS$\uparrow$& PSNR$\uparrow$ & SSIM$\uparrow$ & LPIPS$\downarrow$ & FPS$\uparrow$& PSNR$\uparrow$ & SSIM$\uparrow$ & LPIPS$\downarrow$ & FPS$\uparrow$ & PSNR$\uparrow$ & SSIM$\uparrow$ & LPIPS$\downarrow$ & FPS$\uparrow$ & PSNR$\uparrow$ & SSIM$\uparrow$ & LPIPS$\downarrow$ & FPS$\uparrow$ & PSNR$\uparrow$ & SSIM$\uparrow$ & LPIPS$\downarrow$ & FPS$\uparrow$\\
\midrule
3DGS/CityGS & 27.35 & 0.815	& 0.218 & 298 & 27.44 &	0.866 & 0.205 & 106 &22.03 & 0.814 & 0.213 & 126 & 21.96 & 0.784 & 0.243 &	95 & 25.59	& 0.811	& 0.233	& 195 & 25.13 & 0.802	& 0.295	& 90 \\ 
fast\_gauss & 26.99	& 0.803	& 0.220	& 72 & 27.06 & 0.859 & 0.206 & 30 & 21.73 & 0.804 & 0.216 & 33 & 21.23 & 0.763 & 0.251 & 23 & 24.47	& 0.792	& 0.249	& 39 & 24.02 & 0.792 & 0.293 & 21\\
web-splat &  27.19 & 0.811 & 0.219 & 933 & failed & failed & failed & failed & 21.57 & 0.800 & 0.221 & 404 & 20.83 & 0.732	& 0.270 & 365 & 24.06 & 0.780 & 0.248 & 508 & 23.60 & 0.786	 & 0.302 & 372 \\
TemporalGS & 26.57 & 0.782 & 0.250 & 485 & 26.57 & 0.834 & 0.219 & 236 & 20.90 & 0.743 & 0.255 & 550 & 21.12 & 0.710 & 0.293 & 609 & 24.15	& 0.705	& 0.309	& 753 & 24.54 & 0.763 & 0.337 & 310 \\
fast\_gauss-TGS& 26.52 &0.784 &0.229 &138 & 26.52 & 0.838 &	0.216 & 168       & 20.93 & 0.751 & 0.240 & 143 & 20.74	& 0.717	& 0.267	& 125 & 23.86 & 0.737 & 0.268 & 153 & 23.80 & 0.780	& 0.302 & 132 \\
web-splat-TGS  & 26.71 &0.796 &0.222 &949 & failed & failed & failed & failed & 21.45 & 0.780 &	0.234 & 418 & 20.60 & 0.712	& 0.281	& 374 & 23.71 & 0.748 & 0.269 & 511 & 23.15	& 0.762	& 0.314	& 551 \\
\bottomrule
\end{tabular}}
\end{table*}

\begin{table}[th!]
\Huge
\caption{Quantitative comparison on the Mip-NeRF-360 dataset. Mem means Peak Memory (GB).} 
\label{tab:mipnerf}
\centering
\resizebox{0.4\textwidth}{!}{
\begin{tabular}{l|l|ccccc}
\toprule
Category & Methods &
Mem & PSNR$\uparrow$ & SSIM$\uparrow$ & LPIPS$\downarrow$ & FPS$\uparrow$\\
\midrule
Baseline & 3DGS  & 2.04 &  27.35&0.815&0.218&	298 \\ 
\midrule
\multirow{3}{*}{Post-training} & LightGS & 0.71 &	26.99 & 0.802&	0.239&	377\\
& CP3DGS & 0.75 & 26.21& 0.788&	0.250	&370\\
& StochasticSplats & 1.93 & 24.66 & 0.645 & 0.357 & 184 \\
\midrule 
Training & C3DGS & 0.74 & 26.97&0.800	&0.244	&383\\
\midrule
Post-processing & RadSplat$^{\dagger}$ & 2.17 &  23.63 & 0.753 & 0.264 & 422 \\
\midrule
Temporal-coherence & NeoG & 6.15 & 27.28 & 0.813	& 0.219 & 331
\\
\midrule
\multirow{6}{*}{\makecell[l]{Training-free\\plug-and-play}} & 3DGS-TGS &  2.15 & 26.83&0.795	& 0.226	& 372 	\\
& LightGS-TGS & 0.77 & 26.48 &0.780&0.248&	486\\
& CP3DGS-TGS & 0.85 & 25.83&0.768&0.259&	491 \\
& StochasticSplats-TGS & 2.04 &  24.67 & 0.649 & 0.352 & 194 \\
& C3DGS-TGS & 0.79 & 26.57 &0.782&0.250&485\\
& RadSplat$^\dagger$-TGS & 2.28 & 23.31 & 0.735 & 0.272 & 475 \\
\bottomrule
\end{tabular}}
\end{table}

\subsection{Implementation Details}
We implement our TemporalGS in CUDA and plug it into 3DGS methods without any alteration of their design and implementation. For more implementation details, please refer to the supplementary materials. We conduct all experiments in Python under a headless rendering setting on a single Nvidia RTX 4090 GPU card. We set $\lambda_{{\textup{TDC}}}$ and $\lambda_{{\textup{SR}}}$ to 50 and 5, respectively. Note that we downscale the TGB by a factor of 2 on the large-scale scenes, while we do not downscale the TGB on the small-scale scenes.

\subsection{Results}
\subsubsection{Comparison to State-of-the-Art Methods}
We compare our TemporalGS with vanilla 3DGS, state-of-the-art 3DGS pruning and compression approaches, i.e., LightGaussian (LightGS)~\cite{fan2024lightgaussian}, CP3DGS~\cite{niedermayr2024compressed}, and C3DGS~\cite{lee2024compact}, post-processing-based RadSplat~\cite{niemeyer2025radsplat}, StochasticSplats~\cite{kheradmand2025stochasticsplats}, and the temporal-coherence-based NeoG~\cite{oh2026neo}. We reproduce LightGS, CP3DGS, C3DGS, StochasticSplats, and NeoG according to their official public available code. We perform the visibility computation on the training images of each dataset to reimplement RadSplat’s visibility-based filtering on vanilla 3DGS, denoted as RadSplat$^\dagger$. We denote our TemporalGS-based extensions as 3DGS-TGS, LightGS-TGS, CP3DGS-TGS, C3DGS-TGS, RadSplat$^\dagger$-TGS and StochasticSplats-TGS respectively. 

In Table~\ref{tab:mipnerf} and \ref{tab:large_scale}, we present quantitative comparisons on the small-scale Mip-NeRF-360 dataset and five large-scale aerial and street-view scene datasets. We use CityGaussian (CityGS)~\cite{liu2024citygaussian} to reconstruct these five large-scale scenes because of their better reconstruction quality. Note that LightGS, CP3DGS, StochasticSplats, and RadSplat can be applied to any reconstructed scenes for post-training or post-processing. Therefore, for a fair comparison, LightGS, CP3DGS, StochasticSplats, RadSplat$^\dagger$, and NeoG share the same scene reconstruction checkpoints as baselines.

For small-scale scenes, as shown in Table~\ref{tab:mipnerf}, compared to methods that require (post-)training, 3DGS-TGS achieves higher PSNR and FPS than CP3DGS and StochasticSplats, and obtains performance on par with LightGS and C3DGS. Specifically, compared to LightGS, 3DGS-TGS is only 5 FPS slower with 0.16 lower PSNR. In addition, it offers a more favorable balance between visual quality and efficiency than post-processing-based RadSplat. Furthermore, 3DGS-TGS, LightGS-TGS, CP3DGS-TGS, and C3DGS-TGS achieve average FPS gains of 74, 109, 121, and 102, equivalent to 25\%, 29\%, 33\% and 27\% faster than their original counterparts, respectively. Meanwhile, it maintains competitive quantitative performance with only 0.52, 0.51, 0.38 and 0.4 drop in PSNR. 

For large-scale scenes, as shown in Table~\ref{tab:large_scale}, CityGS-TGS on average delivers better qualitative results than LightGS, CP3DGS, and StochasticSplats, achieves performance comparable to RadSplat, and provides a better quality-efficiency trade-off than C3DGS. Specifically, compared to LightGS, CityGS-TGS is 20 FPS faster with 0.92 higher PSNR in Table~\ref{tab:large_scale}. As a plug-in algorithm, in Table~\ref{tab:large_scale}, TemporalGS achieves 135\%, 109\%, 148\%, 101\%, 98\%, and 81\% faster than CityGS, LightGS, CP3DGS, StochasticSplats, C3DGS, and RadSplat at a cost of 0.74, 0.45, 0.48, -0.2, 0.19, and 0.72 drop in PSNR, respectively. The results demonstrate not only the competitive rendering performance of TemporalGS but also its generalization capability. 

Furthermore, compared to NeoG, a concurrent temporal-coherence-based work, TemporalGS delivers higher FPS while preserving decent rendering quality. However, NeoG does not scale well to large scenes, as shown in Table~\ref{tab:large_scale}. Although NeoG uses the same checkpoints as our 3DGS-TGS/CityGS-TGS, it requires up to 2.01$\times$ the memory of 3DGS/CityGS, whereas our method incurs only a 0.05$\times$ memory overhead. The presented statistics of Peak Memory (GB) show that the memory consumption of TemporalGS is affordable. Besides, NeoG is a reuse-and-update sorting scheme for 3DGS, which prevents its applicability to sorting-free Gaussian Splatting.

It can be observed that TemporalGS is less efficient on small-scale scenes. One reason is that, compared to large-scale scenes, the percentage of Gaussians participating in each frame rendering is much higher because of less occlusion and more complete scene observation in small-scale scenes. As a result, not many Gaussians can be filtered out through temporal dynamic culling. 

We present visual results in Fig.~\ref{fig:teaser} and additional comparisons, including those based on real user roaming trajectories, in the supplement. It can be seen that TemporalGS can maintain reasonable visual quality. We also present additional quantitative results on OccluScene3D and runtime breakdown comparison in the supplement.

\subsubsection{Comparison/Integration to Hardware Rasterization Methods}
Recently, hardware rasterization has been incorporated into 3DGS rendering through APIs such as OpenGL, Vulkan, and WebGPU. In Table~\ref{tab:hardware}, we compare CUDA-based 3DGS/CityGS and our TemporalGS with two representative hardware-accelerated methods, including web-splat~\cite{niedermayr2024compressed} and fast-gaussian-rasterization (fast\_gauss)~\cite{xu2024representing}. The results show that TemporalGS is comparable to hardware rasterization-based 3DGS methods and can achieve a reasonable balance between rendering quality and acceleration, especially on large-scale scenes. In contrast, the performance advantage of hardware rasterization diminishes as the number of Gaussians increases. This degradation arises from (1) representing Gaussian ellipsoids by quads in traditional hardware rasterization introduces redundancies and (2) a large number of Gaussians lead to many small triangles, which lowers the pixel-to-point ratio and is not well optimized by hardware. Taking into account the above, we implement TemporalGS in CUDA for its popularity and excellent performance across small- and large-scale scenes. Furthermore, to show the portability of our TemporalGS, we adapt our algorithm to fast\_gauss and web-splat, denoted as fast\_gauss-TGS and web-splat-TGS, as shown in Table~\ref{tab:hardware}. The results demonstrate that TemporalGS is compatible with hardware rasterization pipelines and leads to a decent speed-up. It is observed that TemporalGS does not significantly boost web-splat in some cases. The main reason is that depth rendering requires float16 instead of uint8, and WebGPU does not optimize float16 as it does for uint8.

\begin{table}[tp!]
\normalsize
 \caption{Impact of individual modules of TemporalGS.}
\label{tab:ablation}
\centering
    \resizebox{0.8\columnwidth}{!}{
    \begin{tabular}{ccc|cc|cc}
    \toprule
     \multirow{2}{*}{AFC}&\multirow{2}{*}{TOC}&\multirow{2}{*}{SR}  & \multicolumn{2}{c|}{Kitchen} &\multicolumn{2}{c}{Residence}\\
      &&& PSNR$\uparrow$ & FPS$\uparrow$ &PSNR$\uparrow$ & FPS$\uparrow$ \\
      \midrule 
      &\checkmark&\checkmark&30.03&  328& 21.13 & 300\\
      \checkmark&&\checkmark& 30.03 & 313  &  21.16& 258 \\
      &&\checkmark& 30.04& 311 & 21.13 & 246 \\
      \checkmark&\checkmark&& 30.82 & 244 & 22.05 & 165  \\
      \checkmark&\checkmark&\checkmark& 30.03 & 331 & 21.16&  320 \\
      \midrule
      \multicolumn{3}{c|}{3DGS/CityGS} & 30.84 & 231 & 22.03 & 126   \\
      \bottomrule 
    \end{tabular}}
 
\end{table}

\begin{table}[tp!]
\normalsize
\caption{Impact of camera motion speeds. We simulate different camera motion speeds using different interpolation intervals. SR Tile(\%) denotes the average percentage of \#SR rendered tiles among the total number of tiles in each SR frame.}
\label{tab:interp}
\begin{minipage}{0.49\columnwidth}
\centering
\resizebox{\linewidth}{!}{
    \begin{tabular}{c|ccc}
    \toprule
    \multirow{2}{*}{Interval} & \multicolumn{3}{c}{Kitchen} \\
     & PSNR$\uparrow$ & FPS$\uparrow$ & SR Tile(\%) \\
    \midrule
      0.02& 30.03 & 331 & 33.59  \\
      0.05& 30.15 &282 & 45.65 \\
      0.1 & 30.29 & 255 &63.87 \\
    \bottomrule
    \end{tabular}
}
\end{minipage}
\hfill
\begin{minipage}{0.49\columnwidth}
\centering
\resizebox{\linewidth}{!}{
    \begin{tabular}{c|ccc}
    \toprule
    \multirow{2}{*}{Interval} & \multicolumn{3}{c}{Residence} \\
     & PSNR$\uparrow$ & FPS$\uparrow$ & SR Tile(\%) \\
    \midrule
    0.1 & 21.16 & 320 & 9.22 \\
    0.2 & 21.58	& 302 & 15.84 \\
    0.4 & 21.23	& 260 & 23.69 \\
    \bottomrule
    \end{tabular}
}
\end{minipage}
\end{table}

\subsubsection{Ablation Study}
We conduct ablation studies on representative scenes, i.e., the Kitchen scene from the small-scale scene dataset and the Residence scene from the large-scale scene datasets.

\noindent\textbf{Impact of Individual Modules} We adopt a leave-one-out strategy to evaluate three key modules of TemporalGS, i.e., AFC, TOC, and SR. The results are shown in Table~\ref{tab:ablation}. For high FPS rendering, it is not surprising that the SR module plays the most significant role in TemporalGS, followed by TOC and AFC. TemporalGS gains 87/155 FPS with SR, 18/62 FPS with TOC and 3/20 FPS with AFC on Kitchen/Residence scenes, respectively. For image quality, TOC and AFC have a small impact on PSNR. 

\noindent\textbf{Impact of Camera Motion Speeds} To investigate the robustness of TemporalGS to camera motion speeds, similar to NeRFBuff~\cite{liu2024nerfbuff}, we simulate different camera motion speeds by using different interpolation intervals on test trajectory generation. A bigger interval corresponds to a faster speed. The assumption behind TemporalGS is that neighboring frames share similar geometry and appearance. Therefore, rapid camera movements may cause FPS performance degradation due to less content overlap. From Table~\ref{tab:interp}, it can be seen that the FPS decreases as the interval increases. This can be explained by a corresponding increase in the percentage of SR rendered tiles, denoted as SR Tile(\%). Even so, TemporalGS still shows a considerable gain in FPS when the camera moves fast. Note that the PSNR remains similar as the interval changes. Also, since the test trajectory of Kitchen exhibits less content overlap between consecutive frames than Residence, it has a higher percentage of SR rendered tiles.

\begin{table}[tp!]
 \caption{Impact of different hyperparameters of TemporalGS. TGSF(\%) denotes the percentage of \#TemporalGS frames (rendered with AFC+TOC+SR) among the total number of rendered frames.}
\label{tab:hyperpara}
\normalsize
\centering
    \resizebox{0.8\columnwidth}{!}{
    \begin{tabular}{ccc|ccc|ccc}
    \toprule 
    \multirow{2}{*}{$\lambda_{{\textup{TDC}}}$}&\multirow{2}{*}{$\lambda_{{\textup{SR}}}$} & \multirow{2}{*}{$\eta (\%)$} & \multicolumn{3}{c|}{Kitchen} &\multicolumn{3}{c}{Residence}\\
      &&& PSNR$\uparrow$ & FPS$\uparrow$ & TGSF(\%) &PSNR$\uparrow$ & FPS$\uparrow$& TGSF(\%) \\
      \midrule
      50& 5& 20 & 30.171& 305& 46.98& 21.156 & 315 & 79.08\\
      50& 5& 40 & 30.026 & 331 & 62.74 & 21.156 & 320 & 80.00\\
      50& 5& 60 & 29.907 & 349 & 76.08 & 21.156 & 320 & 80.00\\
      \midrule
      75& 5& 40& 30.021 & 331 & 62.96 & 21.159 & 320 & 80.00\\
      50& 5& 40& 30.026 & 331 & 62.74 & 21.156 & 320 & 80.00\\
      25& 5& 40& 30.034 & 331 & 62.62 & 21.166 & 315 & 80.00\\
      \midrule
      50& 10& 40& 30.029 & 341 & 68.35 & 21.084 & 335 & 89.44\\
      50& 5& 40& 30.026 &  331 & 62.74 & 21.156 & 320 & 80.00\\
      50& 2& 40& 30.060 &  324 & 49.96 & 21.711 & 241 & 50.00\\
      \bottomrule 
    \end{tabular}}
\end{table}

\noindent\textbf{Impact of Hyperparameters} 
The key hyperparameters of TemporalGS include the TDC frame interval $\lambda_\textup{TDC}$, the SR frame interval $\lambda_\textup{SR}$, and the maximum allowed percentage of SR rendered tiles $\eta$. From Table~\ref{tab:hyperpara}, it can be seen that TemporalGS is most sensitive to changes of $\eta$ because $\eta$ directly controls the number of TemporalGS frames (rendered with AFC+TOC+SR) through the adaptive reference frame insertion mechanism. In the Residence scene, the results change slightly as $\eta$ increases from 20 to 60. This is because, as shown in Table~\ref{tab:interp}, its percentage of SR rendered tiles is 9.22\%, which is much smaller than the minimal $\eta$ value, 20. Therefore, the proposed adaptive mechanism is rarely triggered to insert reference frames. In addition, this mechanism makes TemporalGS less sensitive to $\lambda_\textup{SR}$ and $\lambda_\textup{TDC}$. When these two parameters become large, the percentage of TemporalGS frames (rendered with AFC+TOC+SR), denoted as TGSF(\%), is bounded, and a good trade-off in rendering quality and speed is maintained. 

\section{Conclusion, Limitation, and Future Work}
In this paper, we present TemporalGS, the first training-free plug-and-play algorithm for 3DGS rendering acceleration. By leveraging novel temporal priors and interleaving two acceleration strategies, TemporalGS speeds up rendering without post-training or post-processing and can be easily integrated into existing 3DGS systems. Extensive experiments show that TemporalGS achieves performance comparable to or better than existing state-of-the-art post-training or post-processing 3DGS rendering acceleration methods. TemporalGS can serve as a drop-in module to significantly boost the rendering speed of various 3DGS methods across software- and hardware-based 3DGS rasterization. The major limitation is that the frame similarity assumption might not hold when the camera moves too fast. However, it is rare in real user roaming scenarios. Besides, when the adaptive threshold $\eta$ is set too high, too few reference frames are inserted, and the accumulated errors can result in flickering effects. For future work, it is promising to further enhance rendering quality by introducing a lightweight network to remove artifacts caused by TemporalGS. It is also interesting to develop a TemporalGS counterpart in 4D Gaussian Splatting.

\bibliographystyle{IEEEtran}
\bibliography{sample-base}

\end{document}